%% file: main.tex
\documentclass[10pt,twocolumn,letterpaper]{article}

\usepackage{cvpr}              
\input{preamble}
\definecolor{cvprblue}{rgb}{0.21,0.49,0.74}
\usepackage[pagebackref,breaklinks,colorlinks,allcolors=cvprblue]{hyperref}

\newcommand{\fref}[1]{Fig.~\ref{#1}}
\newcommand{\tref}[1]{Tab.~\ref{#1}}
\usepackage[table]{xcolor}
\usepackage{stfloats}

\def\paperID{8156} 
\def\confName{CVPR}
\def\confYear{2026}

\title{Domain-Shift Immunity in Deep Deformable Registration via Local Feature Representations}

\author{Mingzhen Shao\\
Kahlert School of Computing \\
Scientific Computing and Imaging Institute\\
University of Utah\\
{\tt\small shao@cs.utah.edu}
\and
Sarang Joshi\\
Department of Biomedical Engineering\\
Scientific Computing and Imaging Institute\\
University of Utah\\
{\tt\small sarang.joshi@utah.edu}
}

\begin{document}
\maketitle
\input{sec/0_abstract}    
\input{sec/1_intro}

\input{sec/2_related_work}
\input{sec/3_method}

\input{sec/4_experiment}
\input{sec/5_conclusion}
{
    \small
    \bibliographystyle{ieeenat_fullname}
    \bibliography{main}
}

\input{sec/X_suppl}

\end{document}

%% file: sec/0_abstract.tex
\begin{abstract}

Deep learning has advanced deformable image registration, surpassing traditional optimization-based methods in both accuracy and efficiency. However, learning-based models are widely believed to be sensitive to domain shift, with robustness typically pursued through large and diverse training datasets—without explaining the underlying mechanisms.
In this work, we show that \textbf{domain-shift immunity is an inherent property} of deep deformable registration models, arising from their reliance on \textbf{local feature representations} rather than global appearance for deformation estimation. To isolate and validate this mechanism, we introduce UniReg, a universal registration framework that decouples feature extraction from deformation estimation using fixed, pre-trained feature extractors and a U-Net–based deformation network.
Despite training on a single dataset, UniReg exhibits robust cross-domain and multi-modal performance comparable to optimization-based methods. Our analysis further reveals that failures of conventional CNN-based models under modality shift originate from dataset-induced biases in early convolutional layers. These findings identify local feature consistency as the key driver of robustness in learning-based deformable registration and motivate backbone designs that preserve domain-invariant local features.

\end{abstract}


%% file: sec/1_intro.tex
\section{Introduction}
\label{sec:intro}

Image registration is a fundamental component in computer vision pipelines, particularly when aligning data acquired at different time points, from different subjects, or across heterogeneous imaging modalities.
Broadly, registration can be categorized into rigid and deformable (non-rigid) alignment. Rigid registration applies a global geometric transformation, whereas deformable registration estimates spatially varying local transformations, enabling alignment of structures undergoing non-linear shape changes.

Recent advances in deep learning have substantially impacted image registration. A wide range of learning-based models have been proposed for both rigid and deformable tasks~\cite{detone2016deep, nguyen2017unsupervised, zhang2019oanet, hoffmann2022synth, Tian_2022_arXiv, tian2024unigradicon}, offering improved accuracy and efficiency over traditional optimization methods.
However, a central question remains: \textbf{Do deep registration models remain robust under domain shift?}

Unlike recognition tasks that operate within a specific image distribution, registration inherently requires cross-domain robustness. Classical optimization-based methods rarely struggled with this issue because their mechanisms were deterministic and input-agnostic. In contrast, learning-based models depend on the statistics of the training domain, and their robustness under domain shift remains poorly understood. Prior attempts to improve generalization typically focus on training with large and diverse datasets, without explaining the underlying mechanisms that govern robustness.

A recent study~\cite{shao2024analyzing} showed that deep rigid registration models can exhibit domain-shift immunity, largely because deformation estimation is driven by local features rather than global appearance.
This finding raises the central question of this paper:
Does the same structure-inherent immunity extend to deep deformable registration models, despite their increased complexity?

In this work, we investigate domain-shift robustness in deep deformable registration. While deformable registration is widely used in medical imaging, our goal is to study generalization, not domain-specific performance. To avoid the restrictions of medical-only datasets, we evaluate models across diverse domains—including natural, facial, and medical images—while maintaining controlled deformation behavior using synthetically generated deformation fields. We focus on 2D registration for simplicity and broader domain coverage, as the underlying deformation mechanisms extend naturally to 3D.


Before proceeding, we clarify our use of the term domain-shift immunity.
For deterministic, non–learning-based methods such as SyN~\cite{avants2008sym}, registration accuracy may vary across domains; however, such variation does not imply a loss of immunity. What matters is that the method remains functionally effective—that is, capable of producing meaningful, anatomically plausible deformation fields regardless of the input domain.
Under this definition, domain-shift immunity does not require identical accuracy across domains. Differences in accuracy can always be improved through architectural choices, regularization strategies, or similarity metrics, and are not the focus of this study.
Our goal is instead to determine whether deep deformable models retain the ability to perform valid registration under domain shifts, similar to classical methods.


Through extensive experiments across multiple network architectures and diverse test domains, we show that deep deformable registration models exhibit the same structure-inherent domain-shift immunity observed in the rigid setting.
To investigate the origin of this immunity, we introduce a universal registration model (UniReg) based on the hypothesis that deformation estimation relies primarily on local feature representations.

UniReg employs fixed, pre-trained feature extractors to generate consistent local features across inputs, which are then processed by a U-Net–based deformation estimator. Freezing the extractor ensures that the network learns solely from local features. The comparable accuracy between UniReg and a conventional U-Net trained directly on raw images supports the conclusion that local features—not image appearance—govern deformation field estimation.

Our multi-modal experiments further reveal that failures in classic CNN-based models do not stem from the deformation architecture itself but from dataset-induced biases in early convolutional layers, which prevent consistent local feature extraction across modalities.

Our main contributions are summarized as follows:

\begin{itemize}
    \item 
    We demonstrate that structure-inherent domain-shift immunity extends to deep deformable registration, and that this robustness originates from the reliance on local feature representations.
    
    \item 
    We identify the failure mode of conventional CNN-based models in multi-modal tasks: early layers overfit to single-domain appearance and fail to extract consistent local features under modality shift.
   
    \item 
    We propose UniReg, a universal deformable registration model that decouples feature extraction from deformation estimation by using fixed, pre-trained feature extractors, resulting in robustness that is far less dependent on training dataset diversity.
  
    %

\end{itemize}

%% file: sec/2_related_work.tex
\section{Related work}
\label{sec:related_work}

\subsection{Deep learning registration}
 Early learning-based registration approaches focused on supervised training using ground-truth deformation fields~\cite{cao2017deformable, rohe2017svf, sokooti2017nonrigid, yang2017quicksilver}. These ground-truth warps were typically obtained either by (1) simulating synthetic deformations or (2) using classical optimization-based algorithms to pre-register paired images. Although supervised methods demonstrated the feasibility of learning deformation mappings, their performance and generalization capacity were fundamentally constrained by the quality and diversity of the available ground-truth deformation fields.

 To overcome these limitations, the field has largely shifted toward unsupervised (self-supervised) deformable registration, in which the model learns directly from image pairs without requiring ground-truth deformation fields. Representative works include De~\etal~\cite{de2017end, de2019deep}, Balakrishnan~\etal~\cite{balakrishnan2019voxelmorph}, Hoffmann~\etal~\cite{hoffmann2022synth}, and Tian~\etal.~\cite{Tian_2022_arXiv} have shown that unsupervised methods can outperform classical optimization-based registration in both accuracy and computational efficiency, and have become the dominant paradigm in medical image registration.

In most unsupervised frameworks, the network predicts a deformation field $\phi$, which is applied to the moving image $I_m$ to produce a warped image. Training typically minimizes a loss of the form:
\begin{equation}
\mathcal{L}(I_f, I_m, \phi) = \mathcal{L}_{\mathrm{Sim}}(I_f, I_m \circ \phi) + \lambda \mathcal{L}_{\mathrm{Smooth}}(\phi),
\end{equation}

where $\mathcal{L}_{\mathrm{Sim}}$ measures image similarity between the fixed image $I_f$ and the warped image $I_m \circ \phi$, and $\mathcal{L}_{\mathrm{smooth}}$ regularizes the deformation to encourage spatial smoothness.
The balance parameter $\lambda$ controls the trade-off between data fidelity and deformation regularity.

A large body of work has focused on designing or improving the similarity term, as it directly supervises the deformation learning process.
For mono-modal registration, where corresponding structures share consistent intensity patterns, simple photometric losses—such as Mean Squared Error (MSE) or Normalized Cross-Correlation (NCC)—have proven effective~\cite{chen2025survey}.
However, these intensity-based metrics break down under multi-modal registration, where intensity relationships differ across modalities (e.g., MRI T1 vs. T2, CT vs. MRI), prompting research into modality-robust similarity measures and feature-based methods.

\subsection{Multi-modal registration}
 Multi-modal registration remains a challenging problem because corresponding anatomical structures may exhibit substantially different intensity patterns—or even appear or disappear—across imaging modalities. Classical optimization-based registration methods typically address this issue using modality-invariant similarity measures, most notably Mutual Information (MI)~\cite{viola1997alignment, wells1996multi}. MI became widely adopted because it captures statistical dependence between images without relying on direct intensity correspondence.

 Before MI and related measures were fully incorporated into deep learning–based registration, early neural approaches often circumvented the multi-modal challenge by relying on paired multi-modal datasets or image synthesis strategies. For example, Cao~\etal.~\cite{cao2018deep} exploited pre-aligned CT–MR training pairs to convert the difficult multi-modal similarity problem into a simpler mono-modal one: the network learns alignment by comparing the warped moving image to its paired counterpart in the same modality. This approach avoids defining a multi-modal similarity metric directly but requires paired, co-registered datasets that are rarely available in practice.

 More recent work has explored incorporating modality-robust similarity losses directly into deep learning frameworks, including MI~\cite{guo2019multi}, correlation ratio~\cite{roche1998correlation, chen2025correlation}, and structural or descriptor-based measures such as MIND~\cite{heinrich2012mind}, cross-modal descriptors~\cite{chen2015using}, and handcrafted multi-modal features~\cite{chen2017cross}.
However, integrating these losses into end-to-end learning is nontrivial: MI and correlation ratio require continuous approximations of intensity distributions (\eg, via Parzen windowing) to enable backpropagation, leading to a trade-off between statistical accuracy, differentiability, and computational cost. Descriptor-based losses mitigate some of these issues but often introduce additional parameters, heuristic feature designs, or modality-specific assumptions.

\subsection{Towards generalizable model}
 Developing registration models that generalize across diverse imaging domains and modalities has been a long-standing goal in the field.
 A common strategy in prior work is to expand the training distribution so that models are exposed to a wider range of anatomical variations, appearance characteristics, and transformation patterns. For example, Tian~\etal.~\cite{tian2024unigradicon} leverage large-scale datasets to improve cross-dataset robustness, while Li~\etal.~\cite{li2025unireg} introduce a conditional control vector that encodes anatomical regions, registration types, and instance-specific cues, enabling task-adaptive deformation prediction.
 Other approaches aim to synthesize additional training diversity: Chen~\etal.~\cite{chen2025pretraining} and Dey~\etal.~\cite{dey2024learning} generate large collections of synthetic images or deformation fields to expose models to broad modality and appearance variations.

 Although these strategies improve robustness by enlarging or enriching the training distribution, they \textbf{treat generalization as a data-coverage problem}. As a result, their performance remains tightly coupled to the diversity of the training data and often degrades when deployed on previously unseen modalities or domains.
 Importantly, these methods do not investigate the intrinsic mechanisms that enable or limit generalization in deformable registration—such as the role of local feature representations, early-layer bias, or the structural properties of deformation estimation.
 Consequently, their ability to generalize is bounded by the distribution of the training data rather than by properties of the registration model itself.


%% file: sec/3_method.tex
\section{Method}
\label{sec:method}

To generalize our analysis, we focus primarily on 2D deformable registration tasks, noting that the key difference between 2D and 3D cases lies only in the dimensionality of their operations. To maintain consistency across diverse domains and even multi-modality scenarios, we generate synthetic image pairs using a deformable transformation matrix. This strategy provides \textbf{controlled and consistent deformation patterns} for evaluation.

\subsection{Data generation}

\subsubsection{Synthetic deformation field}
To generate synthetic training pairs, we construct a smooth deformation field $\boldsymbol{\phi}$ from randomly sampled stationary velocity fields (SVFs). 
Specifically, we generate $N$ independent SVFs $\{\mathbf{v}_n\}_{n=1}^N$, where $N$ corresponds to the number of channels in the input images. 
Each voxel in $\mathbf{v}_n$ is independently drawn from a normal distribution 
$\mathcal{N}(0, \sigma_j^2)$ at a coarse spatial resolution $r_p$, 
where the variance $\sigma_j$ is randomly sampled from a uniform distribution 
$\mathcal{U}(0, b_p)$. 
The hyperparameter $b_p$ controls the overall deformation strength. 
The sampled SVFs are then integrated over time to obtain a smooth diffeomorphic 
deformation field and subsequently upsampled to the full image resolution:
\[
\boldsymbol{\phi} = \exp(\mathbf{v}),
\]
where $\exp(\cdot)$ denotes the integration of the stationary velocity field via the scaling-and-squaring method.

 \subsubsection{Training data (LabelM)}

  We use generalized label maps as training data instead of domain-specific images. Each label map contains $J$ distinct regions represented by random geometric patterns. We first generate $J$ smoothly varying noise images $\{p_j\}_{j=1}^{J}$, where each $p_j$ is sampled from a standard normal distribution $\mathcal{N}(0, 1)$ at a coarse resolution $r_p$ and then upsampled to the target image size. 
  The final label map $s$ is constructed by assigning to each pixel $k$ the label index corresponding to the noise image with the maximum intensity at that pixel:
    \[
    s_k = \arg\max_{j} ([\tilde{p}_j]_k),
    \]
  where $\tilde{p}_j$ denotes the upsampled version of $p_j$. This procedure yields randomized yet spatially coherent label maps that are independent of any specific domain, serving as a generalized training set for the registration network. 

\begin{figure}
      \centering
      \includegraphics[width=0.95\linewidth]{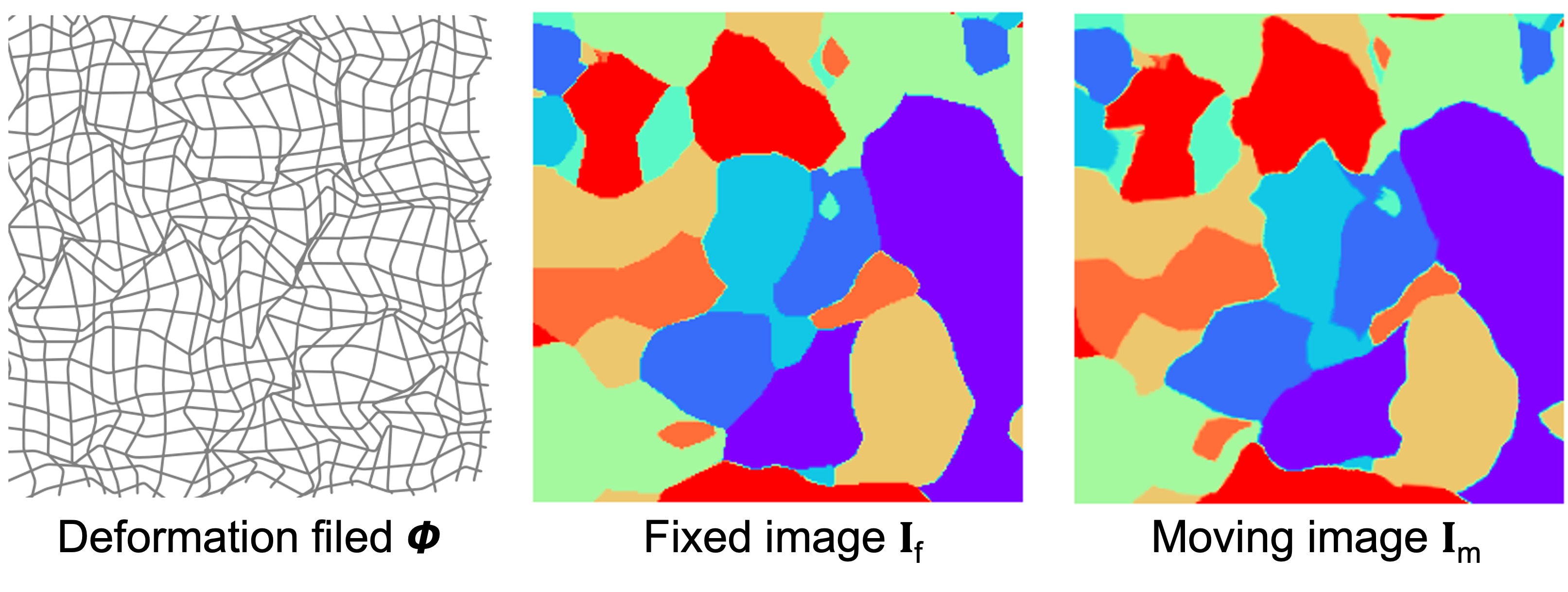}
      \caption{Deformation field and training image pairs.}
      \label{fig:train}
  \end{figure}

  We randomly crop a 224x224 patch from the label map $s$ as the fixed image $I_f$ and applying a synthetic deformation field $\phi$ to $I_f$ produces a transformed map $I_m$. The resulting pairs $(I_f, I_m)$ serve as our training dataset LabelM. \fref{fig:train} illustrates the deformation field $\phi$, the fixed image $I_f$, and the moving image $I_m$.

\subsubsection{Multi-modal data}
 When discussing multi-modal data, medical images such as T1-T2 or MRI-CT pairs often come to mind. However, since our focus is on domain-shift immunity in general image registration tasks, these datasets are relatively narrow in scope and do not provide sufficient diversity. To address this, we generate controlled and consistent multi-modal image pairs using three complementary methods.

\textbf{Multi-modal LabelM:} 
A random recoloring process is applied to the $I_m$ in LabelM to create multi-modal pairs, preserving the underlying spatial correspondences.

\textbf{Multi-modal BSD:} We transform a normalized grayscale image $I \in [0,1]^{H \times W}$ into a color image $I' \in [0,1]^{H \times W \times 3}$ using a continuous, monotonic mapping \[C: [0,1] \rightarrow [0,1]^3,\]
which assigns low-intensity values to dark colors and high-intensity values to bright colors. This transformation preserves the spatial structure of $I$ while altering its photometric distribution, producing a modality-like variation that maintains geometric correspondence across images.

\textbf{T1-T2 Image Pair:} 
To align with standard multi-modal registration tasks, we also include T1–T2 image pairs from the IXI dataset~\footnote{https://brain-development.org/}. We select T1 and T2 volumes acquired from the same scan session to ensure consistent anatomical representation and approximate alignment. From each volume pair, we extract 2D slices at random axial positions to generate multi-modal image pairs.

\begin{figure}[ht]
     \centering
     \includegraphics[width=0.95\linewidth]{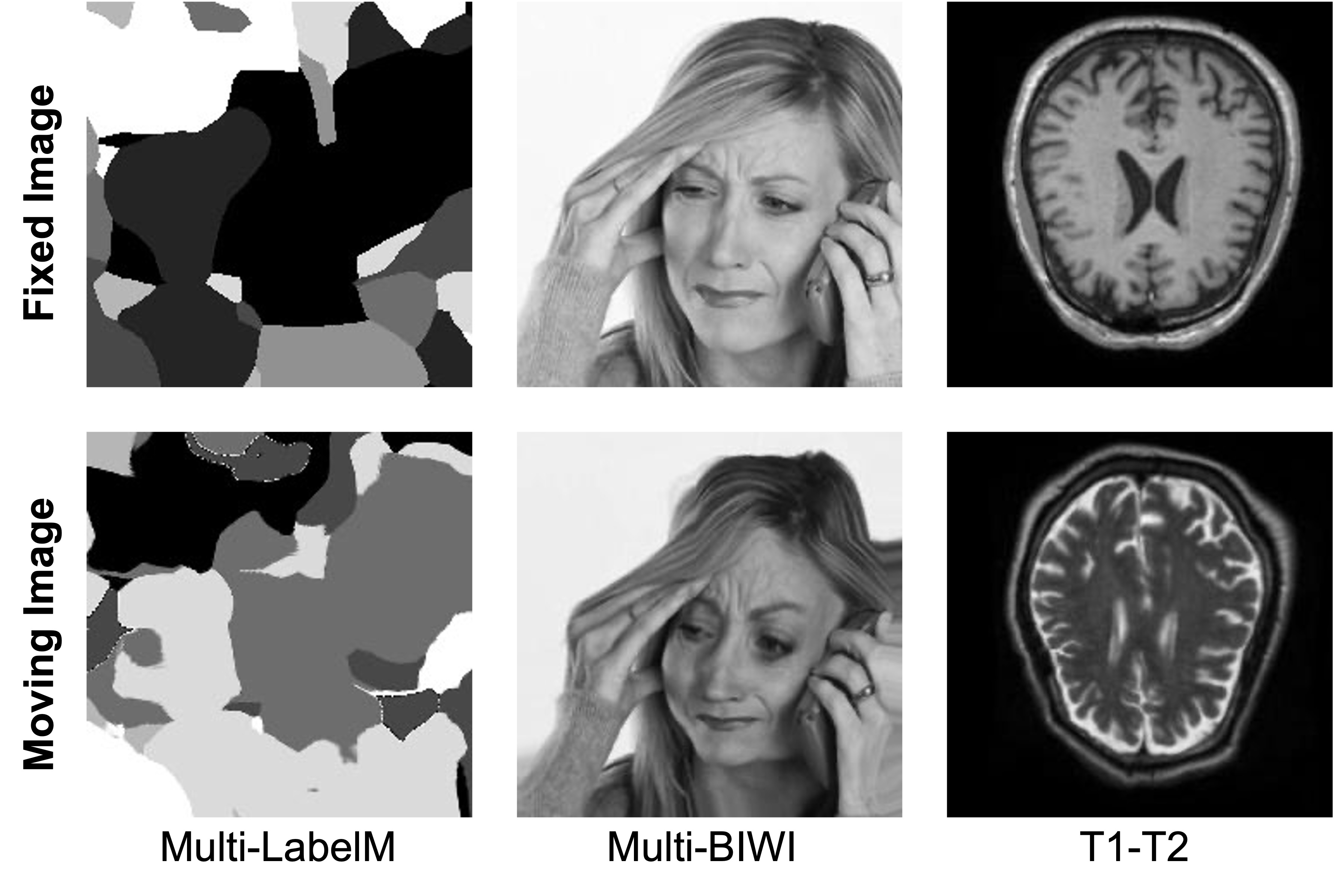}
     \caption{Cross-domain multi-modal examples}
     \label{fig:multi}
 \end{figure}

Figure~\ref{fig:multi} illustrates example image pairs generated using these three methods.

\subsection{Network architecture}
 \begin{figure*}[ht]
     \centering
     \includegraphics[width=0.9\linewidth]{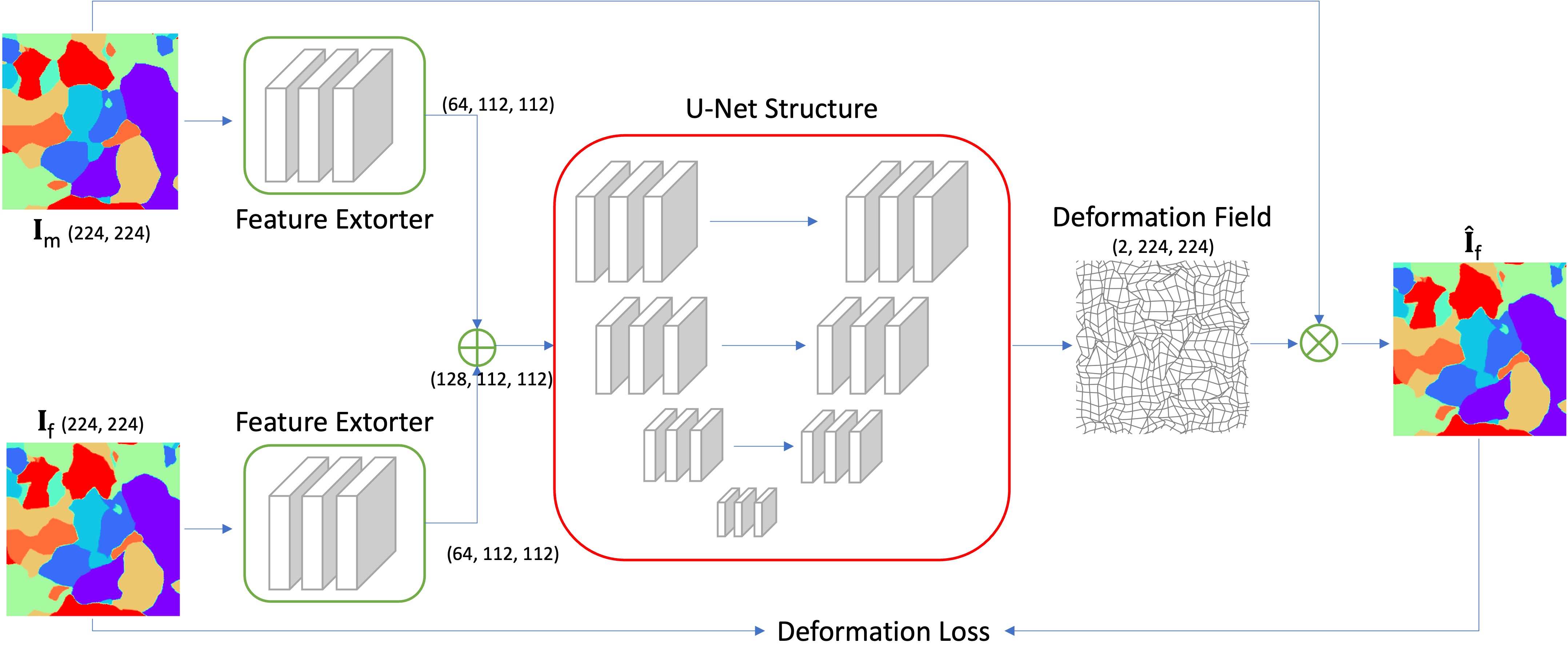}
     \caption{Overview of the proposed universal registration model.}
     \label{fig:network}
 \end{figure*}

 The overview of our universal registration model (UniReg) is illustrated in \fref{fig:network}. The main difference from previous works is that our model import a \textit{Feature Extractor} block to preprocess the fixed image $I_f$ and moving image $I_m$, and then concatenates the extracted local feature fed to a U-Net architecture. The U-Net architecture prediction a deformation field $\phi$ and then applied to the $I_m$ to get a new wrapped image $\hat{I}_f = \phi \circ I_m$. A deformation loss is then calculated between the wrapped image $\hat{I}_f$ and the fixed image $I_f$. 

 \subsubsection{Feature extractor}
    We use the layers up to the first convolutional block of VGG16 as our feature extractor. The extractor is initialized with weights pre-trained on ImageNet and remains frozen during training. This design ensures that the extracted tensors emphasize low-level local features (\eg edges, textures) that are less sensitive to domain variations. Both input images share the same feature extractor, and their resulting feature maps are concatenated before being fed into the subsequent registration network.
    
 \subsubsection{FeatureUNet}
    The FeatureUNet follows an encoder–decoder design but is tailored for feature-level image registration. The network takes as input the concatenated feature maps extracted from the feature extractor and processes them through a three-level encoder–decoder pathway. And then an additional upsample layer is applied to make sure the output deformation field is in the same resolution as the input image. 

 \subsubsection{Deformation loss}
    \begin{equation}
        \mathcal{L}_{\mathrm{smooth}} = \frac{1}{N} \sum_{i,j} \Big( \| \nabla_x \mathbf{\phi}_{i,j} \|_1 + \| \nabla_y \mathbf{\phi}_{i,j} \|_1 \Big),
    \end{equation}
    where $\mathbf{\phi}\in\mathbb{R}^{2\times H\times W}$ is the predicted flow field, $N$ is the total number of the deformable field, and $\nabla_x,\nabla_y$ denote the discrete spatial gradient (finite-difference) operators along the vertical and horizontal axes respectively.

    \begin{equation}
     \mathcal{L}_{\mathrm{reg}} = \mathrm{MSE}\!\left(I_f, \hat{I}_f\right) + \alpha\, \mathcal{L}_{\mathrm{smooth}},
    \end{equation}
    where $\mathrm{MSE}$ is the mean square error and $\alpha = 0.01$ is a hyperparameter to adjust the smooth strength.

\subsection{Validation metrics}

Traditional registration evaluation often relies on anatomical correspondence measures such as DICE and mean Target Registration Error (mTRE), which require either segmentation masks or manually annotated landmark pairs. However, these annotations are typically unavailable for natural image datasets and are inconsistent across domains. Since our goal is to evaluate \emph{general-purpose} registration performance that extends beyond medical imaging, we instead adopt \textit{intensity-based similarity metrics} that do not rely on ground-truth anatomical labels.

\textbf{Correlation Coefficient (CC):}
For {mono-modal} registration, we measure similarity using the CC to capture linear intensity agreement between the fixed image $I_f$ and the warped moving image $\hat{I}_f$:
\begin{equation}
\mathrm{CC}(I_f, \hat{I}f) =
\frac{\sum{i} (I_f(i) - \bar{I}_f),(\hat{I}_f(i) - \overline{\hat{I}f})}
{\sqrt{\sum{i}(I_f(i) - \bar{I}f)^2},\sqrt{\sum{i}(\hat{I}_f(i) - \overline{\hat{I}_f})^2}},
\end{equation}
where $\bar{I}_f$ and $\overline{\hat{I}_f}$ denote the mean intensities of $I_f$ and $\hat{I}_f$, respectively.

\textbf{Mutual Information (MI):}
For multi-modal registration, we use MI to measure statistical dependence rather than direct intensity correspondence:
\begin{equation}
\mathrm{MI}(I_f, \hat{I}_f) = H(I_f) + H(\hat{I}_f) - H(I_f, \hat{I}_f),
\end{equation}
where $H(\cdot)$ and $H(\cdot,\cdot)$ denote marginal and joint entropies.


\textbf{Deformation Regularity:}
To assess whether the predicted deformation is smooth and physically plausible, we examine the Jacobian matrix of \(\phi\), \[J_{\phi}(x) = \frac{\partial \phi(x)}{\partial x} ,\]
whose determinant \(\det(J_{\phi}(x))\) measures the local change in area induced by the transformation. 
Regions where \(\det(J_{\phi}(x)) \le 0\) correspond to \emph{folding}, indicating that the mapping is no longer locally invertible. 
We quantify deformation regularity using the percentage of non-positive Jacobians:
\[\%|J| = \frac{\big|\{x \mid \det(J_{\phi}(x)) \le 0\}\big|}{|\Omega|} \times 100\%.\]
Lower \(\%|J|\) indicates smoother and more anatomically consistent deformations.

%% file: sec/4_experiment.tex
\section{Experiments}

In this section, we first analyze the domain-shift immunity of deep deformable registration models. We then use the proposed network to show that this immunity arises from the reliance on local feature representations. Finally, in the multi-modal setting, we demonstrate that incorporating a fixed, pre-trained feature extractor in our UniReg allows the model to better exploit this immunity, resulting in substantially improved robustness and performance.

To ensure generality, we focus on the fundamental architectural components of deep learning–based registration models rather than specific medical registration frameworks such as VoxelMorph or GradICON. Although these models differ in training strategies, they largely share the same U-Net backbone and therefore offer limited insight across different network structures.

The \colorbox{gray!15}{shaded column} in tables indicates that all models were trained on this domain; the remaining columns evaluate cross-domain generalization.

\subsection{Domain-shift immunity}

 \begin{figure*}[ht]
     \centering
     \includegraphics[width=0.95\linewidth]{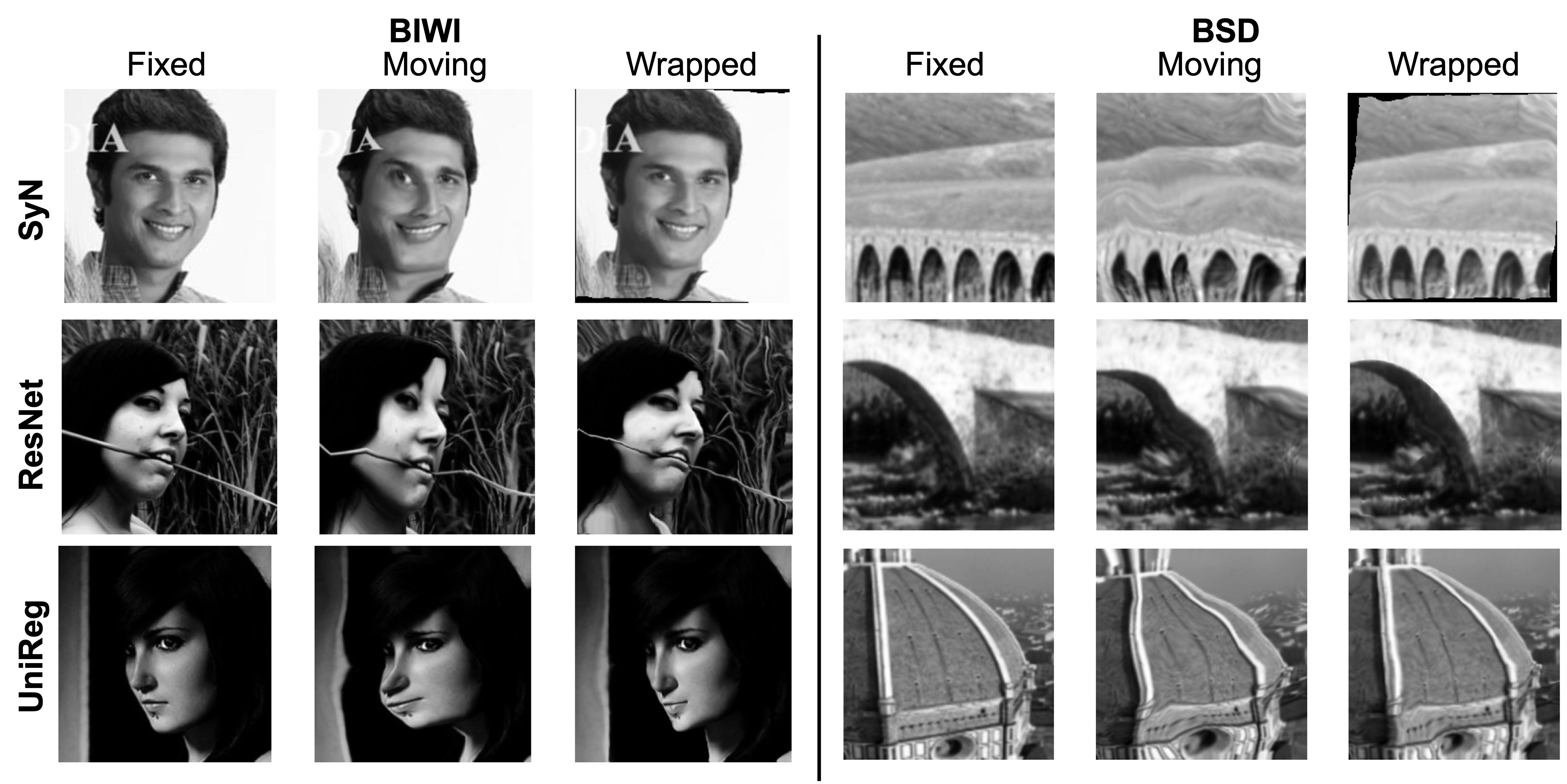}
     \caption{Example mono-modal registration from different models.}
     \label{fig:cross}
 \end{figure*}

To verify the existence of domain-shift immunity, we follow the same evaluation strategy as in~\cite{shao2024analyzing}, where a model is trained on one domain and directly tested on unseen domains without any additional fine-tuning or modification. We include a non-learning registration method (SyN) as a baseline. 
As a well-established non-learning registration algorithm, SyN does not rely on domain-specific training and is thus inherently domain-independent.
If a deep learning model outperforms this baseline on an unseen domain, it indicates that the model remains functional under domain shifts.  

To ensure that our findings are not architecture-specific, we evaluate multiple backbone networks, including VGG16, ResNet50, and U-Net. To achieve diverse dataset coverage, we employ BSD300 (natural images), BIWI (facial images), and NIH chest X-ray~\cite{Wang2017chest} (X-ray images), which collectively span a broad range of visual characteristics and application domains. For architectures such as VGG16 and ResNet50, we append a series of upsampling layers after the final convolutional block to ensure that the output deformation field matches the resolution of the input image.

\begin{table}[ht]
\setlength{\tabcolsep}{4pt}
\centering
\caption{Registration results across domains. Values are reported as CC ($\%|J|$), where CC ($\uparrow$) and $\%|J|$ ($\downarrow$).}
\begin{tabular}{l >{\columncolor{gray!15}}c c c c}
\hline
             & LabelM & BIWI & BSD & X-Ray \\
\hline
SyN          & 0.92(0) & 0.89(0) & 0.82(0) & 0.86(0) \\
\cline{1-1}
VGG16          & 0.97(0.82) & 0.95(0.47) & 0.93(0.37) & 0.96(0.11) \\
Res50        & 0.97(1.13) & 0.95(0.76) & 0.94(0.42) & 0.97(0.19) \\
U-Net        & 0.98(1.24) & 0.96(0.92) & 0.93(0.54) & 0.96(0.13) \\
\hline
\end{tabular}
\label{tab:domain}
\end{table}

        


Table~\ref{tab:domain} reports the correlation coefficients and $\%|J|$ across the four evaluation domains. All deep learning–based methods achieve low folding rates while consistently outperforming SyN in appearance similarity, indicating that learned registration models generalize effectively to unseen data. 
Moreover, the consistent performance trends observed across different network architectures demonstrate that domain-shift immunity is a general property of learning-based deformable registration.



 \textbf{Domain-shift immunity \& local feature: }
 Unlike rigid registration, deep deformable registration relies on more complex network architectures, making it difficult to directly analyze the contribution of local features. However, given the similarity in task objectives and domain-shift immunity across models, we hypothesize that this robustness arises from their shared reliance on local features when estimating the deformation field. Motivated by this, we design UniReg to explicitly isolate and test this mechanism, and we conduct the following experiments to validate our hypothesis.


\subsection{Universal registration model}
 We categorize domain-shift scenarios into two settings: (1) mono-modal registration, where both input images belong to the same modality, and (2) multi-modal registration, where the inputs originate from different modalities.
 Across all experiments, we show that by using a fixed feature extractor, UniReg substantially reduces dependence on training data diversity and delivers robust cross-domain generalization.


 \subsubsection{Mono-modal registration}

 The mono-modal registration follows the same experimental setup as described in Sec.~4.1; therefore, we only report the results of our UniReg here.

 \begin{table}[ht]
 \setlength{\tabcolsep}{4pt}
     \centering
     \caption{Mono-modal registration results of the UniReg across domains. Values are reported as CC ($\%|J|$).}
     \begin{tabular}{l >{\columncolor{gray!15}}c c c c}
     \hline
        Models    &   LabelM  &  BIWI  &  BSD   & X-ray     \\        
    \hline
         UniReg &  0.98(1.38)  &  0.96(0.96)  &  0.94(0.48)  &  0.97(0.19) \\
         \hline
     \end{tabular}
     \label{tab:mono}
 \end{table}
 By comparing \tref{tab:domain} and \tref{tab:mono}, we observe that the universal registration model achieves accuracy comparable to that of U-Net. This similarity can be partly attributed to their shared encoder–decoder architecture.
 
 However, it is important to note that in our model, the U-Net structure in UniReg is trained \textbf{purely using local features} extracted from a fixed, pre-trained feature extractor. In contrast, the original U-Net has direct access to the full intensity information from raw input images. Despite this limitation, our model still provide a similar accuracy, indicating that local features alone play a critical role in accurate deformation field estimation.
 \fref{fig:cross} shows example registration results for several models. A complete set of results for all models is provided in the supplementary material.


 \subsubsection{Multi-modal registration}
 We begin by training all models on the mono-modal LabelM dataset, following the setup in Sec.~4.1. For reference, we compute the initial MI between each fixed–moving image pair before registration. The overall performance is summarized in ~\tref{tab:multi}.

  
 \begin{table}[ht]
     \centering
     \caption{Multi-modal registration results (trained on mono-modal). Values are reported as MI ($\%|J|$). MI ($\uparrow$) and $\%|J|$ ($\downarrow$). }
     \begin{tabular}{lcccc}
     \hline
        Models  &  Multi-LabelM  & Multi-BIWI  &  T1-T2  \\
    \hline
         Init MI &     1.17    &    0.80    &    0.51         \\
         \cline{1-1}
         VGG16  &    0.86(12.69)   &   0.98(0.79)    &  0.59(5.02)             \\
         Res50   &  0.85(18.14)   &   0.98(1.20)   &  0.61(6.31)               \\
         U-Net &    0.89(21.15)  &  0.99(1.60)   &   0.67(6.52)          \\
         \hline
         UniReg  &   0.87(24.58)  &   1.05(1.55)   &  0.69(8.48)           \\
    \hline
     \end{tabular}
     \label{tab:multi}
 \end{table}

 We first observe that performance on Multi-BIWI is substantially higher than on Multi-LabelM and T1–T2, even though its folding rate ($\%|J|$) is nearly 10× higher than typical values in previous mono-modal registration, indicating that the deformation field overfits to local intensity patterns rather than meaningful geometric alignment. This disparity can be attributed to dataset characteristics: while all three are multi-modal, Multi-BIWI exhibits only a mild modality shift, as its thermal images preserve much of the structural content of the intensity domain (see \fref{fig:cross}). In contrast, Multi-LabelM and T1–T2 introduce larger modality gaps. Consequently, Multi-BIWI behaves more like a mono-modal registration task, resulting in higher MI despite degraded regularity.

 To probe the cause of these failures, we retrain all models using two multi-modal datasets with varying appearance diversity—Multi-LabelM (high diversity) and Multi-BIWI (low diversity)—under identical training protocols. The predicted deformation field is applied to the mono-modal image, and the same loss function is used for optimization.
 
  \begin{table}[ht]
     \centering
     \caption{Multi-modal registration results (trained on Multi-LabelM). Values are reported as MI ($\%|J|$).}
     \begin{tabular}{l >{\columncolor{gray!15}}c c c}
     \hline
        Models  &  Multi-LabelM  & Multi-BIWI  &  T1-T2  \\
    \hline
         Init MI &     1.17    &    0.80    &    0.51         \\
        \cline{1-1}
         VGG16  &   1.64(0.15)   &   1.10(0.11)    &   0.66(0.05)                \\
         Res50   &  1.62(0.44)   &   1.17(0.10)  &  0.65(0.11)                \\
         U-Net &    1.74(0.64)  &  1.10(0.11)   &   0.65(0.18)          \\
         \hline
         UniReg  &   1.69(0.50)  &   1.05 (0.07)   &    0.66(0.04 )         \\
    \hline
     \end{tabular}
     \label{tab:multi_train}
 \end{table}

   \begin{table}[ht]
   \setlength{\tabcolsep}{4pt}
     \centering
     \caption{Mono-modal registration results (trained on Multi-LabelM). Values are reported as CC ($\%|J|$).}
     \begin{tabular}{l c c c c}
     \hline
        Models  &  LabelM  & BIWI  &  BSD  &  X-ray  \\
    \hline
         VGG16  &   0.97(0.13)   &   0.95(0.06)    &   0.92(0.07)   &   0.98(0.05)          \\
         Res50   &  0.97(0.35)   &   0.95(0.08)  &  0.92(0.04)    &   0.97(0.03)      \\
         U-Net &    0.99(0.28)  &  0.95(0.02)   &   0.92(0.01)  &  0.98(0.01)          \\
         \hline
         UniReg  &  0.99(0.43)   &   0.95 (0.05)   &    0.92(0.01)   &   0.97(0.02)      \\
    \hline
     \end{tabular}
     \label{tab:multi_train_mono}
 \end{table}

We report both multi-modal and mono-modal registration results trained on Multi-LabelM in ~\tref{tab:multi_train} and \ref{tab:multi_train_mono}. After training on Multi-LabelM, all models recover stable cross-domain performance and achieve improved mono-modal accuracy with lower folding rates.

For conventional architectures (VGG16, ResNet50, U-Net), this improvement stems from better feature generalization in the early convolutional layers. When trained solely on mono-modal LabelM, these layers overfit to single-domain intensity distributions and fail to extract domain-invariant local features.
Training on Multi-LabelM mitigates this overfitting, leading to robust performance across modalities.

UniReg also benefits: with a fixed extractor, multi-modal inputs still produce useful local features, but the registration head can overfit; training on Multi-LabelM regularizes this head and stabilizes deformation prediction.

  \begin{table}[ht]
     \centering
     \caption{Multi-modal registration results (trained on Multi-BIWI). Values are reported as MI ($\%|J|$).}
     \begin{tabular}{l c >{\columncolor{gray!15}}c c}
     \hline
        Models  &  Multi-LabelM  & Multi-BIWI  &  T1-T2  \\
    \hline
         Init MI &     1.17    &    0.80    &    0.51         \\
         \cline{1-1}
         VGG16  &   1.18(0.00)   &   0.82(0.00)    &   0.50(0.00)            \\
         Res50   &  1.18(0.00)   &   0.82(0.00)  &  0.51(0.00)          \\
         U-Net &    0.84(22.97)  &  1.28(0.43)   &   0.76(7.83)          \\
         UniReg* &   1.18(13.61)   &   0.82(0.12)    &   0.52(4.1)             \\
         \hline
         UniReg  &   1.78(0.72)  &   0.99 (0.05)   &    0.63(0.03 )         \\
    \hline
     \end{tabular}
     \label{tab:multi_train_biwi}
 \end{table}

To validate this observation, we repeat the same experiment using Multi-BIWI, a dataset with substantially lower appearance diversity (\tref{tab:multi_train_biwi}).
Under this setting, VGG16 and ResNet50 fail to learn meaningful deformations, yielding near-initial MI values and degenerate Jacobian regularity. U-Net performs well only on the source domain but struggles to extract transferable features for unseen modalities.

In contrast, UniReg remains functional, while its fine-tuned variant (UniReg*), which adapts the feature extractor, collapses in the same way as conventional models.

\begin{figure}[ht]
     \centering
     \includegraphics[width=0.95\linewidth]{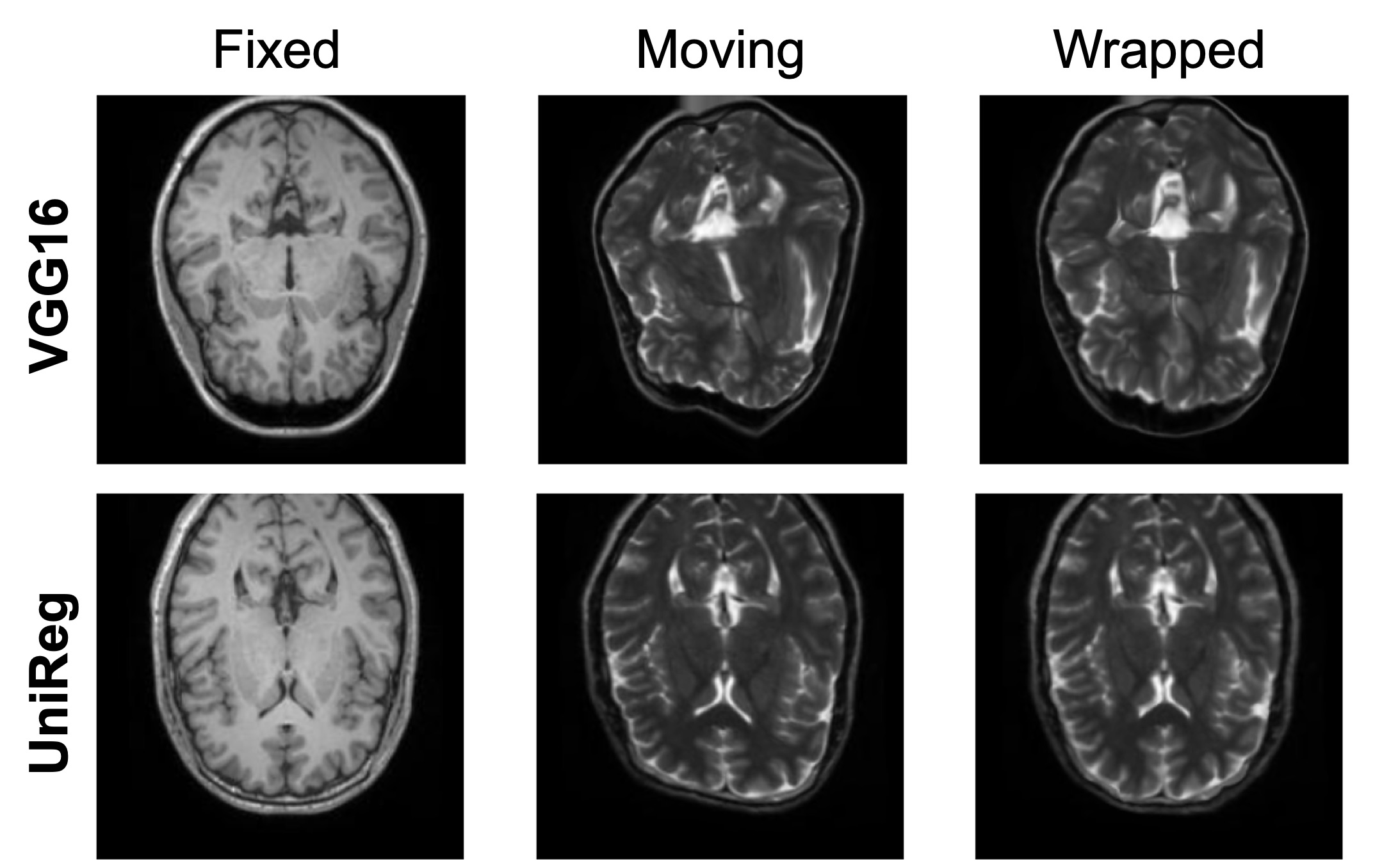}
     \caption{Example multi-modal registration from different models.}
     \label{fig:multi}
 \end{figure}

This comparison highlights that conventional models fail to generalize across domains when their feature extractors (the early convolutional layers) are trained on low-diversity data and cannot produce consistent local representations. In contrast, UniReg avoids this failure mode: its fixed, pre-trained feature extractor provides stable local features across modalities, enabling reliable cross-domain performance even under limited training diversity.
\fref{fig:multi} shows example multi-modal registration results trained on Multi-LabelM for several models. A complete set of results for all models is provided in the supplementary material.

\subsection{Discussion}
Our experiments demonstrate that deep deformable registration models possess structure-inherent domain-shift immunity, as deformation estimation relies primarily on local feature representations rather than raw intensity appearance.
The multi-modal results further show that robustness is not inherently lost when modalities differ; instead, degradation occurs only when the feature extractor fails to produce consistent local representations across domains.

Unlike conventional CNN-based models that jointly learn appearance encoding and deformation prediction, UniReg decouples these stages through a fixed, pre-trained feature extractor.
This design prevents overfitting to the appearance statistics of the training set and maintains stable registration even under limited modality diversity.
While UniReg’s extractor remains static, training with high-diversity multi-modal data still improves robustness by regularizing how the registration network combines local features to infer deformation fields.

These findings highlight that feature extraction—not deformation modeling—is the primary bottleneck for cross-domain generalization.
Although U-Net remains the dominant architecture for deformable registration, our results suggest that architectures leveraging invariant feature encoders (e.g., VGG or other pre-trained extractors) can achieve stronger robustness under unseen domains.
Thus, UniReg serves as a proof-of-concept for a broader design principle: decoupling feature representation from deformation estimation can yield domain-stable registration models regardless of the specific backbone or feature extractor used.

%% file: sec/5_conclusion.tex
\section{Conclusion}


Our study reveals that deep deformable registration models possess structure-inherent domain-shift immunity, driven by their reliance on local feature representations. The analysis with UniReg confirms that maintaining consistent local features is sufficient for robust cross-domain and multi-modal registration, and that failures in conventional CNN-based models largely arise from dataset-induced biases in early layers. These findings shift the focus from deformation architecture to feature extraction as the key determinant of generalization, suggesting that future research should re-examine backbone design and develop feature-extractor architectures explicitly optimized for cross-domain robustness.

%% file: sec/X_suppl.tex
\clearpage
\setcounter{page}{1}
\maketitlesupplementary

\section{Different feature extractor}

 As discussed in the main paper, UniReg is a proof-of-concept framework and either component can be replaced with different architectures without altering the core principle.

To verify this claim, we report additional results using alternative feature extractors.
Tab.~\ref{tab:diff_extractor} and \ref{tab:diff_extractor_multi} present mono-modal and multi-modal registration performance, respectively, for UniReg variants that replace the original extractor with different pretrained networks.
UniReg(Res) uses the early layers of ResNet-50, and UniReg(Goog) uses the early layers of GoogLeNet; in both cases, we retain all layers up to the first pooling operation.
Each extractor is initialized with ImageNet pre-trained weights and remains frozen during training.

Across both mono-modal and multi-modal settings, the choice of feature extractor has minimal impact on the overall performance of UniReg.
These results reinforce the conclusion that universality does not rely on the specific architecture of the feature extractor, but rather on the decoupling itself: once the appearance encoder provides sufficiently informative local features, the deformation network can generalize robustly across domains.

 \begin{table}[ht]
\setlength{\tabcolsep}{3pt}
\centering
\caption{Registration results across domains. Values are reported as CC ($\%|J|$), where CC ($\uparrow$) and $\%|J|$ ($\downarrow$).}
\begin{tabular}{l >{\columncolor{gray!15}}c c c c}
\hline
             & LabelM & BIWI & BSD & X-Ray \\
\hline
UniReg(Res)    & 0.99(1.12) & 0.96(0.86) & 0.94(0.36) & 0.97(0.08) \\
UniReg(Goog)    & 0.98(1.09) & 0.95(0.72) & 0.92(0.43) & 0.94(0.10) \\
\hline
\end{tabular}
\label{tab:diff_extractor}
\end{table}

  \begin{table}[ht]
  \setlength{\tabcolsep}{3pt}
     \centering
     \caption{Multi-modal registration results (trained on Multi-BIWI). Values are reported as MI ($\%|J|$).}
     \begin{tabular}{l c >{\columncolor{gray!15}}c c}
     \hline
        Models  &  Multi-LabelM  & Multi-BIWI  &  T1-T2  \\
    \hline
         Init MI &     1.17    &    0.80    &    0.51         \\
         \cline{1-1}
         UniReg(Res)  &   1.82(0.70)  &   1.02 (0.04)   &    0.65(0.02 )     \\
         UniReg(Goog)  &   1.63(0.78)  &   0.96 (0.05)   &    0.59(0.04 )     \\
    \hline
     \end{tabular}
     \label{tab:diff_extractor_multi}
 \end{table}

\section{Training details using multi-modal datasets}
 In Sec.~4.2.2, we train all models on the Multi-LabelM and Multi-BIWI datasets. The training process is summarized as follows.

 During training data generation, we first construct the multi-modal image $I_{fM}$ from the fixed image $I_f$ using the procedure described in Sec.~3.1.3.
 
 We then apply the synthetic deformation field $\phi$ to both $I_{fM}$ and $I_f$ to obtain the transformed images $I_{mM}$ and $I_m$ respectively. 
 
 The model receives the pair $(I_f, I_{mM})$ as input and predicts the deformation field $\hat{\phi}$. 
 Since $I_f$ and $I_m$ belong to the same modality, the loss is computed between $I_f$ and $\hat{\phi} \circ I_m$, ensuring that supervision is applied in a mono-modal space even though the inputs are multi-modal.

\section{VoxelMorph}
 We also evaluated cross-domain performance using a VoxelMorph architecture. As discussed in Sec.~4, such architectures are specifically designed for medical image registration and are not ideal for analyzing general deformable registration behavior. We report these results here only to verify that the observed domain-shift immunity still holds under this model family.

\begin{table}[ht]
\setlength{\tabcolsep}{3pt}
\centering
\caption{Registration results across domains. Values are reported as CC ($\%|J|$), where CC ($\uparrow$) and $\%|J|$ ($\downarrow$).}
\begin{tabular}{l >{\columncolor{gray!15}}c c c c}
\hline
             & LabelM & BIWI & BSD & X-Ray \\
\hline
VoxelMorph    & 0.99(1.16) & 0.95(0.83) & 0.92(0.48) & 0.96(0.11) \\
\hline
\end{tabular}
\label{tab:voxel}
\end{table}

\section{Supplementary qualitative results on toy datasets}

 \begin{figure}[ht]
     \centering
     \includegraphics[width=0.85\linewidth]{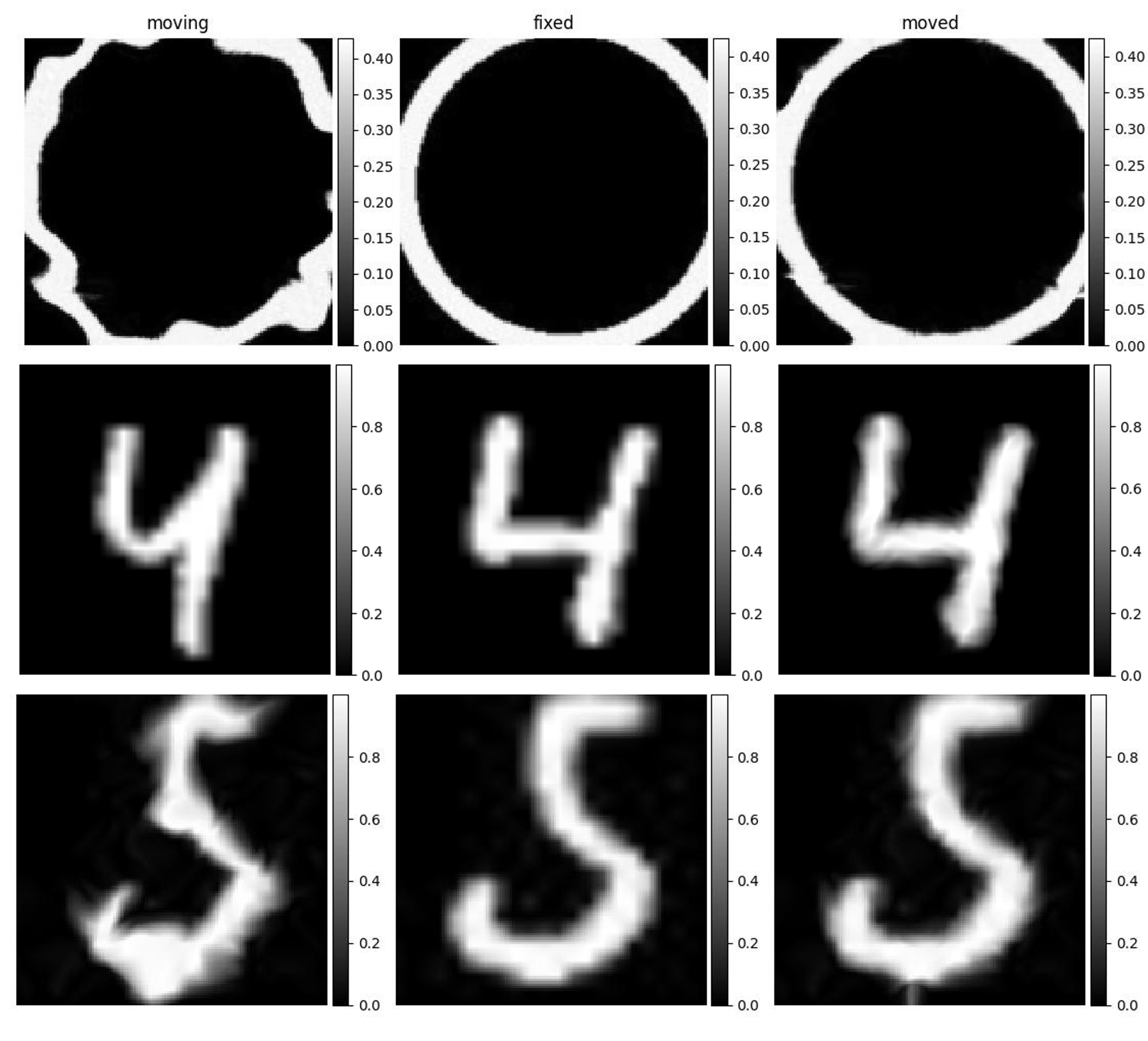}
     \caption{Registration results on MNIST and basic geometric shapes.}
     \label{fig:toy}
 \end{figure}
 
 To further illustrate the generality of our findings, we include additional qualitative results on extremely simple datasets such as MNIST digits and basic geometric shapes (see \fref{fig:toy}). The results are obtained using our UniReg model trained on the LabelM dataset. These datasets produce visually clean and easily interpretable examples of registration behavior and domain-shift robustness. However, because their appearance variability and structural complexity are far lower than those of the datasets used in the main paper, they do not meaningfully challenge the registration models or reflect realistic cross-domain conditions. For this reason, we omit them from the main experiments and present them here solely as supplementary visual evidence supporting our conclusions.

\section{Mono-modal registration}
\begin{figure*}[ht]
    \centering
    \includegraphics[width=0.85\linewidth]{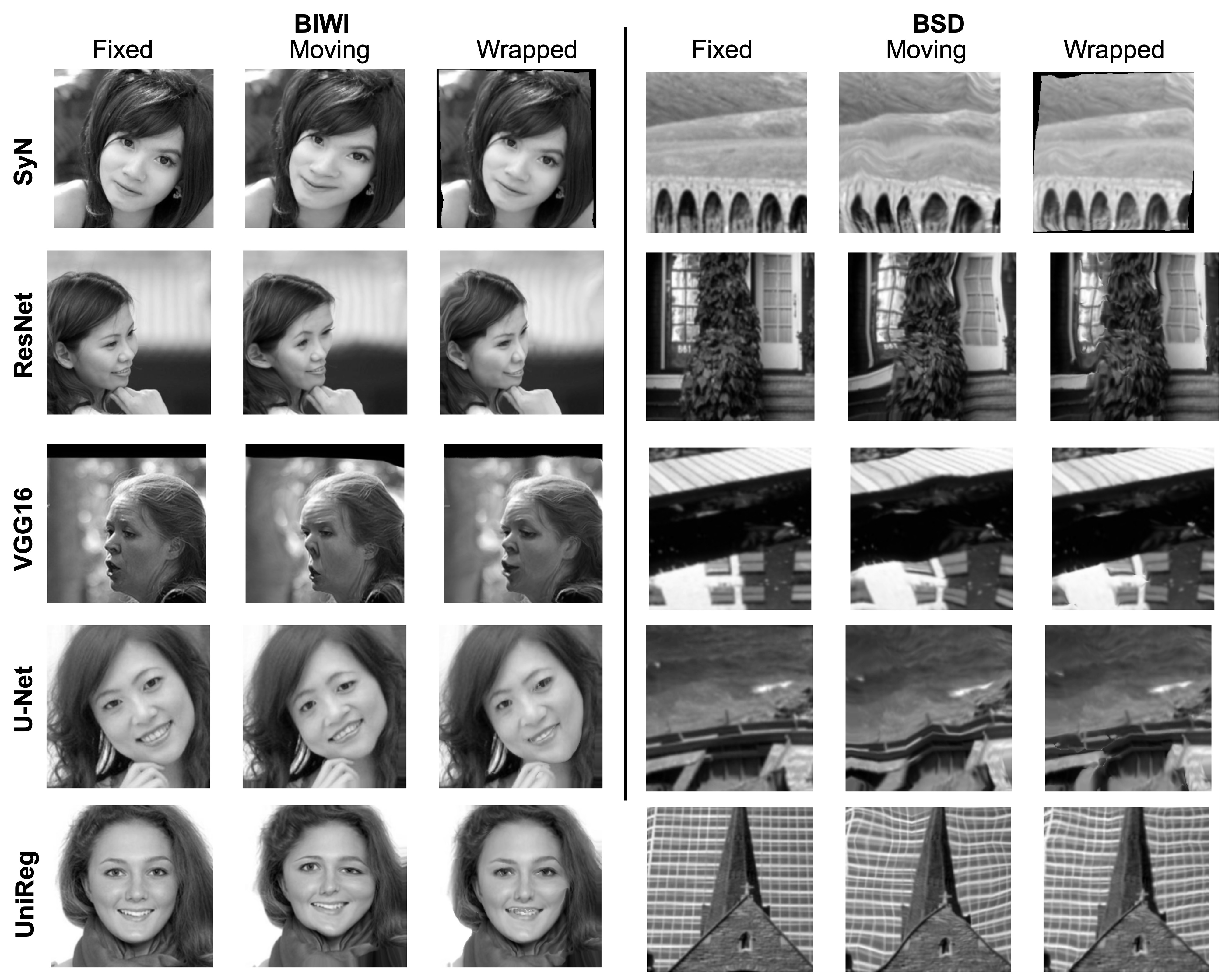}
    \caption{Mono-modal registration from different models.}
    \label{fig:mono_all}
\end{figure*}

Mono-modal registration results for all models in \tref{tab:domain} are provided to illustrate performance, with same-domain examples excluded to reduce redundancy (see \fref{fig:mono_all}). All of the models are trained on LabelM dataset.

\section{Multi-modal registration}
\begin{figure*}[ht]
    \centering
    \includegraphics[width=0.85\linewidth]{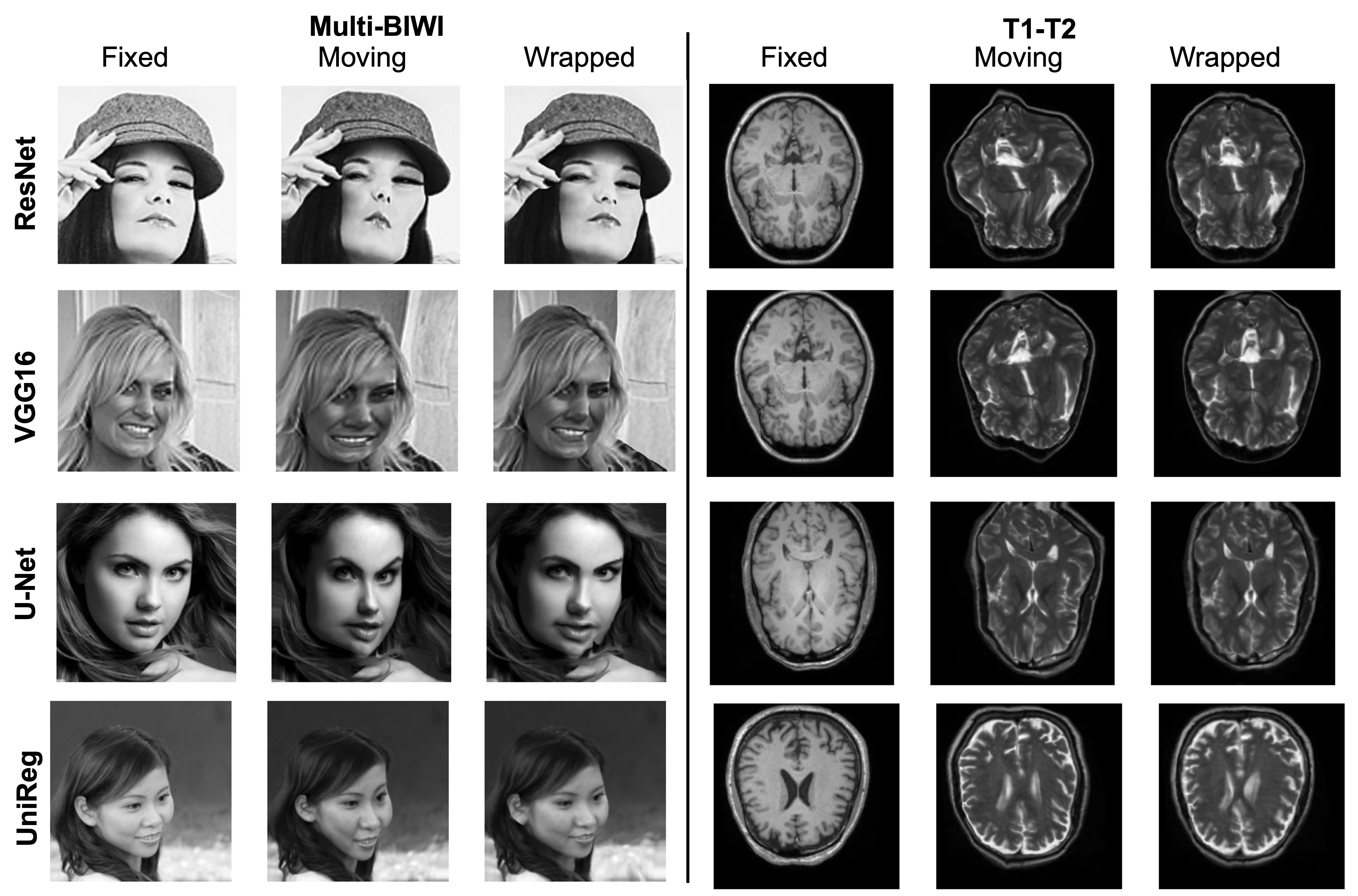}
    \caption{Multi-modal registration from different models.}
    \label{fig:multi_all}
\end{figure*}

Multi-modal registration results trained on Multi-LabelM for all models in \tref{tab:multi_train} are provided to illustrate their performance (see \fref{fig:multi_all}). All of the models are trained on Multi-LabelM dataset.
